\documentclass[conference]{IEEEtran}
\IEEEoverridecommandlockouts
\usepackage{cite}
\usepackage{amsmath,amssymb,amsfonts}
\usepackage{algorithmic}
\usepackage{graphicx}
\usepackage{textcomp}
\usepackage{xcolor}
\usepackage{hyperref}
\usepackage{array}
\usepackage{tabularx}
\usepackage{tikz}
\def\BibTeX{{\rm B\kern-.05em{\sc i\kern-.025em b}\kern-.08em
    T\kern-.1667em\lower.7ex\hbox{E}\kern-.125emX}}
\begin{document}

\title{Measuring What Matters: Benchmarking Generative, Multimodal, and Agentic AI in Healthcare
}

\author{\IEEEauthorblockN{Prasanna Desikan}
\IEEEauthorblockA{
\textit{Centific AI Research}\\
Redmond, WA, USA \\
prasanna.desikan@centific.com}
\and
\IEEEauthorblockN{Harshit Rajgarhia}
\IEEEauthorblockA{
\textit{Centific AI Research}\\
Redmond, WA, USA \\
harshit.rajgarhia@centific.com}
\and
\IEEEauthorblockN{Shivali Dalmia}
\IEEEauthorblockA{
\textit{Centific AI Research}\\
Redmond, WA, USA \\
shivali.dalmia@centific.com}
\and
\IEEEauthorblockN{Ananya Mantravadi}
\IEEEauthorblockA{
\textit{Centific AI Research}\\
Redmond, WA, USA\\
ananya.mantravadi@centific.com}

}

\maketitle

AI models are increasingly deployed in live clinical environments where they must perform reliably across complex, high-stakes workflows that standard training and validation datasets were never designed to capture. Evaluating these systems requires benchmarks: structured combinations of tasks, datasets, and metrics that enable reproducible, comparable measurement of what a model can do. The central challenge in healthcare AI is not performance alone, but the absence of systematic methods to measure reliability, safety, and clinical relevance under real-world conditions. Most existing benchmarks test what a model knows; too few test whether it can perform reliably and without failing across the full complexity of real clinical tasks. Current benchmarks have accumulated through ad hoc dataset construction optimized for narrow task performance: frontier models achieve near-perfect scores on medical licensing examinations, but when evaluated across real clinical tasks, performance degrades sharply, scoring 0.74--0.85 on documentation, 0.61--0.76 on clinical decision support, and only 0.53--0.63 on administrative and workflow tasks \cite{medhelm}. High benchmark scores give a false sense of deployment readiness, and the gap between performance and utility widens precisely as AI systems take on more consequential clinical roles. Without a principled framework for benchmark design, the field cannot determine whether poor clinical performance reflects model limitations or failures in how performance is being measured.


Such degradation is a measurement problem, not purely a model problem, and healthcare AI benchmarking lacks a formal design theory. A systematic meta-evaluation of 53 medical AI benchmarks found that 94\% include no mechanism to test model robustness, 96\% do not evaluate a model’s ability to handle uncertainty, and 92\% do not address data contamination — the condition in which evaluation data was present in a model's training corpus, causing scores to reflect memorization rather than genuine capability \cite{medcheck}. Recent work such as MedThink-Bench has highlighted the need for structured evaluation of expert-level medical reasoning beyond traditional QA benchmarks \cite{medthink}, and broader perspectives have begun to position benchmarking itself as a scientific discipline requiring dynamic, continuously evolving evaluation frameworks \cite{benchmarking_science}. Structured clinician involvement in benchmark construction is essential to ensuring that the capabilities being measured reflect actual clinical reasoning, workflow priorities, and the judgment required in practice. Tasks designed without domain expert input risk optimizing for what is convenient and measurable rather than what is clinically meaningful, producing benchmarks that cannot credibly assess deployment readiness regardless of how well a model scores.

This tutorial addresses that gap directly. We present a structured framework grounded in maturity-aware task design, multi-metric evaluation principles, and a rigorous benchmark engineering lifecycle, intended to guide the development of next-generation healthcare AI benchmarks across care delivery, digital health, and clinical AI environments. It also addresses how clinical subjectivity, particularly in higher-acuity tasks where expert disagreement is inherent, is handled through annotation methodology and metric design rather than treated as noise to be eliminated. 
The scope of this tutorial is limited to clinical and operational AI
evaluation. It does not cover claims-based benchmarking, payer actuarial
modeling, pharmaceutical trial evaluation, drug discovery, or molecular
modeling frameworks, which require distinct validation and regulatory
approaches.

\begin{figure}[!t]
\centering
\begin{tikzpicture}[
    font=\small,
    every node/.style={inner sep=0pt}
]


\fill[blue!10] (0,0) -- (7,0) -- (7,2.0) -- (0,2.0) -- cycle;
\draw[blue!40, line width=0.4pt] (0,0) -- (7,0) -- (7,2.0) -- (0,2.0) -- cycle;

\fill[blue!18] (0,2.0) -- (7,2.0) -- (7,4.0) -- (0,4.0) -- cycle;
\draw[blue!40, line width=0.4pt] (0,2.0) -- (7,2.0) -- (7,4.0) -- (0,4.0) -- cycle;

\fill[blue!28] (0,4.0) -- (7,4.0) -- (7,6.0) -- (0,6.0) -- cycle;
\draw[blue!40, line width=0.4pt] (0,4.0) -- (7,4.0) -- (7,6.0) -- (0,6.0) -- cycle;

\node[align=center] at (3.5,0.5)
    {\footnotesize Clinical notes $\cdot$ Summaries $\cdot$ EHR documentation};
\node[align=center] at (3.5,1.1)
    {\textbf{L1 - Document \& communicate}};

\node[align=center] at (3.5,2.5)
    {\footnotesize Imaging $\cdot$ Audio $\cdot$ Physiological signals};
\node[align=center] at (3.5,3.1)
    {\textbf{L2 - Detect \& interpret}};

\node[align=center] at (3.5,4.5)
    {\footnotesize Triage $\cdot$ Referrals $\cdot$ Care decisions};
\node[align=center] at (3.5,5.1)
    {\textbf{L3 - Act \& coordinate}};

\draw[->, black, line width=0.8pt] (8.2,0) -- (8.2,6.2);
\node[rotate=90, anchor=south] at (8.7,3.1)
    {\footnotesize Increasing benchmark complexity};
\node[anchor=west] at (8.3,0.2) {\footnotesize L1};
\node[anchor=west] at (8.3,6.0) {\footnotesize L3};

\end{tikzpicture}
\caption{L1--L3 maturity taxonomy. Systems progress from documentation 
and communication at L1, through detection and interpretation at L2, to 
active care coordination at L3. Benchmark complexity scales with maturity 
level as clinical risk increases and model reliability decreases.}
\label{fig:maturity}
\end{figure}
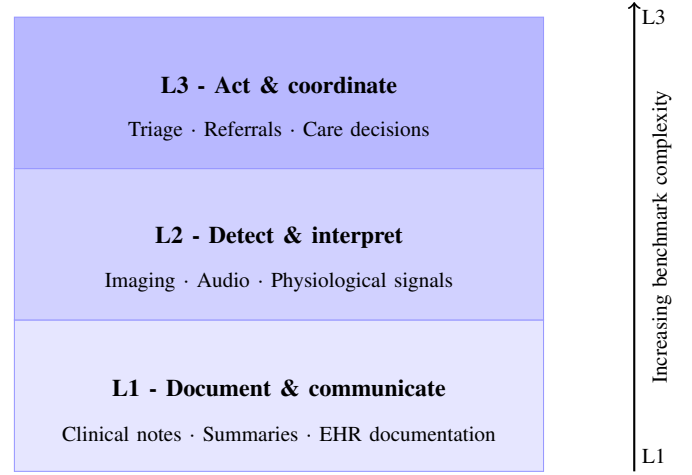

Central to the tutorial is a maturity-based taxonomy reflecting the level of responsibility a system assumes within clinical operations as described in Figure \ref{fig:maturity}. Level 1 (L1): systems that document and communicate, capturing and summarizing clinical information. Level 2 (L2): systems that detect and interpret, identifying clinically relevant signals from structured and unstructured data including imaging, audio, and physiological sensors. Level 3 (L3): systems that act and coordinate, guiding triage, referrals, and care decisions, and in some contexts interacting with device-driven or closed-loop processes. Evidence from recent holistic evaluations of medical language models reveals a consistent gradient: performance is strongest at L1 and weakest at L3, precisely where the stakes
are highest\cite{medhelm, medcheck}. This inverse relationship between task risk and model reliability
makes maturity-aware evaluation not merely a best practice but a clinical necessity.

The tutorial maps evaluation requirements across healthcare AI applications of increasing complexity, spanning clinical information extraction, multimodal medical reasoning, sequential decision-making under uncertainty, and agent-based clinical workflow simulation. Rather than prescribing a fixed taxonomy, the framework identifies the distinct requirements each application class introduces for dataset construction, metric design, annotation, and domain expert validation. Two empirical case studies from the authors' research, a large-scale medical audio reasoning benchmark and an action-based reasoning benchmark for medical AI agents (ART) \cite{art2026} operating within live EHR environments, ground the framework in practice and illustrate benchmark design principles that address the systematic gaps identified. These case studies directly operationalize the three evaluation dimensions the field currently lacks: robustness across task, population, and data integrity variation; uncertainty handling across both model-level and annotation-level ambiguity; and data contamination mitigation through source control, task format novelty, and cross-system generalizability testing.

\begin{table}[t]
\centering
\caption{Tutorial Structure and Content Overview}
\label{tab:tutorial_overview}
\footnotesize
\setlength{\tabcolsep}{3pt}
\renewcommand{\arraystretch}{1.15}

\begin{tabular}{>{\raggedright\arraybackslash}p{0.28\columnwidth}
                >{\raggedright\arraybackslash}p{0.18\columnwidth}
                >{\raggedright\arraybackslash}p{0.46\columnwidth}}
\hline
\textbf{Tutorial Section} & \textbf{Focus} & \textbf{Key Topics} \\
\hline

Opening &
Motivation and problem framing &
Benchmark--utility gap, real-world failures, clinical stakes, tutorial roadmap \\
\hline

Part I -- The Case for Better Healthcare Benchmarking &
Establishing need for better evaluation &
Healthcare AI landscape, modalities (text, image, audio, sensors), benchmarking fundamentals, limits of accuracy, evaluation failures (robustness, uncertainty, temporal reasoning gaps) \\
\hline

Part II -- A Framework for Designing Healthcare Benchmarks &
Core methodology and design principles &
L1--L3 maturity taxonomy, inverse risk--reliability gradient, benchmarking framework, dataset construction, metric design, annotation methodology, clinician validation, five-phase lifecycle \\
\hline

Part III -- Applying the Framework &
Practical application and validation &
Mapping benchmarks (e.g., MedHELM, MedAgentBench), gap detection, ART case study (retrieval, temporal aggregation, conditional logic), deployment readiness, regulatory alignment, go/no-go thresholds \\
\hline

Part IV -- Multimodal Benchmarking in Practice &
Multimodal evaluation deep dive &
Medical audio reasoning benchmark, dataset structure, annotation design, SME involvement, LLM-as-judge, modality bias, performance ceilings, L2 implications \\
\hline

Part V -- The Future of Benchmarking in Healthcare AI &
Forward-looking paradigms &
Dynamic benchmark environments, RL-oriented evaluation, longitudinal journeys, safety stress testing, open problems, call to action \\
\hline

Q\&A &
Discussion and engagement &
Audience questions, deployment challenges, collaboration opportunities \\
\hline

\end{tabular}
\end{table}
Within ART, task variation tests the same clinical reasoning capability across distinct EHR action types such as lab orders versus medication orders. Population variation presents identical clinical conditions across differing patient demographics, including age and sex. Data integrity variation introduces missing labs or vitals as designed stress tests to distinguish models that appropriately express uncertainty from those that hallucinate.

The annotation methodology in the medical audio reasoning benchmark distinguishes between cases where the model lacks sufficient information and should produce a calibrated “I don’t know” response, and cases where the clinical question itself admits no single correct answer. The former is handled through majority-vote annotation, while the latter is resolved through senior clinician adjudication. Contamination is further controlled through de-identified cases from a single health system and cross-system generalizability testing across different regional EHR datasets, where performance degradation would indicate system-specific memorization rather than genuine clinical reasoning.

Table~\ref{tab:tutorial_overview} summarizes the five-part tutorial organization and the major topics covered in each section.

The framework is designed for direct implementation, and the tutorial demonstrates this through the benchmark engineering lifecycle and empirical case studies. The five-phase lifecycle spanning design, dataset construction, technical implementation, validity verification, and governance provides a structured methodology for building and auditing healthcare benchmarks. The two empirical case studies instantiate the framework end-to-end: ART at L3 and the medical audio reasoning benchmark at L2, each walking through task design, dataset construction, and metric selection. Part III applies the framework as a gap-detection tool by auditing existing benchmarks against the proposed design requirements. Full implementation details and design decision walkthroughs will be demonstrated during the tutorial presentation.

Attendees will leave equipped with a practical methodology for translating real healthcare problems into benchmarkable evaluation tasks; designing multimodal datasets including audio and sensor data; selecting metrics that reflect safety and operational impact; applying the medical benchmark engineering lifecycle framework to audit existing or in-development benchmarks; and interpreting benchmark results in regulatory and deployment contexts. 
By grounding evaluation in both maturity level and application domain, this tutorial provides a forward-looking framework for assessing generative models, multimodal agents, and embedded clinical AI systems — and for understanding what current benchmarks cannot yet see.


\section*{Author's Biographies}

\textbf{Prasanna Desikan} is the Head of Healthcare AI Research at Centific. He has held senior leadership roles across healthcare verticals. He has served as Industry Track Chair at ICHI for several years and has led multiple tutorials advancing healthcare AI. 

\textbf{Harshit Rajgarhia} is a Lead Research Scientist at Centific AI Research, where he leads a team building AI systems for major enterprise clients. His research interests include multimodal learning, medical AI reasoning, and reinforcement learning, with publications at VLDB, NeurIPS and ICCV. 

\textbf{Shivali Dalmia} is a Sr. AI Research Solutions Engineer at Centific AI Research, where she built scalable Human-in-the-Loop benchmarking pipelines. Her research spans AI models for clinical decision making and benchmarking, with publications at AAAI, NeurIPS, ICCV, and ICDM. 

\textbf{Ananya Mantravadi} is an AI Research Engineer at Centific AI Research, where her work focuses on reinforcement learning in healthcare, AI evaluation, multi-agent systems. Her publication record includes medical AI agents benchmarking, physiological signal analysis, and multi-agent frameworks, in venues including AAAI, NeurIPS, and the Journal of Systems Architecture.

\newpage


\begin{thebibliography}{9}

\bibitem{medhelm}
Bedi, Suhana, et al. "Medhelm: Holistic evaluation of large language models for medical tasks." arXiv preprint arXiv:2505.23802 (2025).

\bibitem{medcheck}
Ma, Zizhan, et al. "Beyond the leaderboard: Rethinking medical benchmarks for large language models." arXiv preprint arXiv:2508.04325 (2025).

\bibitem{medthink}
S. Zhou \textit{et al.}, “Automating expert-level medical reasoning evaluation of large language models,” 
\textit{Nature Machine Intelligence}, vol. 7, no. 12, pp. 1102--1115, Dec. 2025, doi: 10.1038/s42256-025-07988-x.

\bibitem{benchmarking_science}
M. Z. Ma, M. Saxon, and X. Yue, “The Science of Benchmarking: What’s Measured, What’s Missing, What’s Next,” 
in \textit{Proc. 39th Conf. Neural Information Processing Systems (NeurIPS) Tutorials}, San Diego, CA, USA, Dec. 2025.

\bibitem{art2026}
Mantravadi, A., Dalmia, S., and Mukherji, A., "ART: Action-based Reasoning Task Benchmarking for Medical AI Agents." arXiv preprint arXiv:2601.08988 (2026).


\end{thebibliography}
\end{document}